\documentclass[lettersize,journal]{IEEEtran}

\usepackage{amssymb,amsmath}
\usepackage{cite}
\usepackage{subcaption}
\usepackage{graphicx}
\usepackage{psfrag}
\usepackage{url}
\usepackage[latin1]{inputenc}
\usepackage[absolute,overlay]{textpos}
\usepackage{gensymb}
\usepackage{cases}
\usepackage{physics}
\usepackage{bm}
\usepackage{graphicx}
\usepackage[font=footnotesize]{caption}
\usepackage[font=footnotesize]{subcaption}
\usepackage{float}
\usepackage[linesnumbered,ruled,lined]{algorithm2e}

\usepackage{tabularx}
\usepackage{booktabs}
\usepackage{array}

\allowdisplaybreaks
\usepackage{tikz}
\usepackage{booktabs,multirow}
\usetikzlibrary{patterns,arrows}
\definecolor{color1}{RGB}{237, 191, 193}
\definecolor{color2}{RGB}{229, 153, 157}
\definecolor{color3}{RGB}{225, 123, 116}
\definecolor{color4}{RGB}{10,10,200}
\definecolor{color5}{RGB}{203,52,38}

\usepackage{pgf}

\usepackage[nocomma]{optidef}

\usepackage{algorithmic}

\usepackage{multicol}
\usepackage{amsthm}

\newcommand{\ours}{\text{Fed-Sophia}}

\usepackage{booktabs}
\usepackage{color,soul}

\usepackage[bottom=0.73in,top=0.7in,left=0.57in,right=0.57in]{geometry} \textwidth=7.25in

\usepackage{textcomp}
\usepackage{hyperref}
\usepackage{lipsum}

\newcommand\copyrighttext{%
  \footnotesize \textcopyright 2024 IEEE. Personal use of this material is permitted.
  Permission from IEEE must be obtained for all other uses, in any current or future
  media, including reprinting/republishing this material for advertising or promotional
  purposes, creating new collective works, for resale or redistribution to servers or
  lists, or reuse of any copyrighted component of this work in other works.
  DOI: \href{https://ieeexplore.ieee.org/document/10811936}{10.1109/LWC.2024.3521027}}
\newcommand\copyrightnotice{%
\begin{tikzpicture}[remember picture,overlay]
\node[anchor=south,yshift=10pt] at (current page.south) {\fbox{\parbox{\dimexpr\textwidth-\fboxsep-\fboxrule\relax}{\copyrighttext}}};
\end{tikzpicture}%
}

\begin{document}

\title{Scalable and Resource-Efficient Second-Order Federated Learning via Over-the-Air Aggregation}
\author{
	\IEEEauthorblockN{Abdulmomen Ghalkha, Chaouki Ben Issaid, and Mehdi Bennis\\
	}
	\IEEEauthorblockA{Centre for Wireless Communications (CWC), University of Oulu, Finland}
    }

\maketitle
\copyrightnotice

\begin{abstract}
Second-order federated learning (FL) algorithms offer faster convergence than their first-order counterparts by leveraging curvature information. However, they are hindered by high computational and storage costs, particularly for large-scale models. Furthermore, the communication overhead associated with large models and digital transmission exacerbates these challenges, causing communication bottlenecks. In this work, we propose a scalable second-order FL algorithm using a sparse Hessian estimate and leveraging over-the-air aggregation, making it feasible for larger models. Our simulation results demonstrate more than $67\%$ of communication resources and energy savings compared to other first and second-order baselines.
\end{abstract}

%\vspace{-10pt}

\let\thefootnote\relax\footnote{This work is funded by the European Union under Grant Agreement 101096379. Views and opinions expressed are however those of the author(s) only and do not necessarily reflect those of the European Union or the European Commission (granting authority). Neither the European Union nor the granting authority can be held responsible for them.}
\section{Introduction}
In recent years, the rapid development of the Internet of Things (IoT) and edge computing applications has greatly increased the connectivity requirements of numerous devices and sensors, resulting in massive data generation crucial for training machine learning (ML) algorithms across various applications, including image classification \cite{he2016deep}, speech recognition \cite{chen2020non}, and wireless resource allocation \cite{gao2021machine}. Traditionally, training these ML algorithms involves a centralized approach where devices send their raw data to a parameter server (PS). However, due to privacy constraints, sharing this data may not be possible. Additionally, bandwidth limitations pose a significant challenge when transferring large volumes of data to the PS. Consequently, FL algorithms have become increasingly popular as they enable distributed model training, eliminating the need to share raw and private data \cite{kairouz2021advances}.

Several works have been proposed to enhance privacy and reduce the communication bottleneck in FL. First-order FL algorithms, such as federated averaging (FedAvg) \cite{mcmahan2017communication}, rely on locally computed gradients for model updates, which are aggregated and broadcast by the PS. While computationally less intensive, first-order methods typically suffer from slow convergence. To improve their efficiency, techniques such as momentum and adaptive learning rates have been introduced to accelerate convergence, while quantization and compression schemes are employed to minimize communication overhead \cite{kairouz2021advances}, making first-order methods more practical for resource-constrained environments. However, gradients solely provide a limited view of the objective landscape, resulting in many iterations and communication rounds, necessitating the use of curvature information to ensure faster convergence.

Second-order algorithms were introduced to leverage the Hessian matrix to provide refined update steps, leading to faster convergence. However, despite reducing the number of communication rounds, these methods incur higher computational and communication costs due to the need to compute and transmit the full Hessian matrix. Privacy concerns also arise, as the Hessian contains sensitive information about the local objective function and data. In \cite{ghalkha2024din}, the authors address these issues by sharing the inverse Hessian-gradient product vector. However, these approaches fail to scale up as they require computing and storing the full Hessian, leading to computational complexity and storage requirements of $\mathcal{O}(d^{3})$ and $\mathcal{O}(d^{2})$, respectively, where $d$ is the model size. In \cite{dinh2022done}, the inverse of the Hessian is approximated iteratively and recursively multiplied with the gradient to reduce storage needs. Nevertheless, calculating the Hessian-gradient product still entails a complexity of $\mathcal{O}(d^{3})$, making it inefficient for energy-constrained IoT applications. In \cite{elbakaryfed}, the authors estimated the Hessian as a diagonal matrix by leveraging the optimizer introduced in \cite{liu2023sophia}. However, the models were updated by performing multiple local iterations before aggregating them, which not only imposed a significant computational burden but also introduced the risk of model drift \cite{li2020federated}.

As model size increases, sharing model updates requires substantial bandwidth, making communication a bottleneck. Analog over-the-air (OTA) aggregation, inspired by the superposition property of signals in multiple access channels (MAC), has emerged as a promising technique in first-order methods to reduce communication overhead \cite{nazer2007computation}. However, existing second-order methods have not utilized OTA due to computational and storage costs. The only work exploring OTA and second-order techniques \cite{krouka2022communication} still requires computing and storing the full Hessian.

To leverage the benefits of OTA and accelerate convergence in second-order methods, we introduce OTA Fed-Sophia, an approach that scales up with model size, uses the optimizer from \cite{liu2023sophia}, and reduces bandwidth requirements via OTA aggregation. Our main contributions are summarized as follows
\begin{itemize} 
\item Our approach uses local curvature, avoiding the full Hessian's computation, storage, and sharing, thus reducing computation and communication costs while preserving privacy.
\item To our knowledge, this is the first work to use OTA aggregation with a scalable second-order FL algorithm via channel inversion. 
\item Numerical results show that our approach outperforms baselines with fewer communication resources and lower energy consumption, even for large models. 
\end{itemize}

\section{System Model and Problem Formulation}\label{System_Model}
In the context of FL, we consider $N$ clients that communicate with a PS to learn a model $\bm{\theta}\in\mathbb{R}^d$ that solves the following minimization problem
\begin{align}
    \label{min_problem1}
    \min_{\bm{\theta}\in \mathbb{R}^{d}} \left\{f(\bm{\theta}) \triangleq \sum_{n=1}^{N}f_n(\bm{\theta})\right\},
\end{align}
where $f_n(\bm{\theta})$ is the local loss function of client $n$. 

This formulation aims to minimize a global loss function without necessitating the clients to share their local datasets.
One way to solve the problem in \eqref{min_problem1} iteratively is by utilizing the gradients received from all clients and updating the model in the opposite direction of the calculated aggregate using an appropriate learning rate $\eta$ as follows
\begin{align}
    \label{first_order_update}
    \bm{\theta}^{k+1} = \bm{\theta}^{k} - \eta \nabla f(\bm{\theta}^{k}) = \bm{\theta}^{k} - \eta\sum_{n=1}^{N} \nabla f_n(\bm{\theta}^{k}),
\end{align}

where $\nabla f(\bm{\theta}^{k})$ is the gradient at the $k$th round.

Alternatively, instead of using a uniform step size, clients can utilize second-order information, i.e., the Hessian matrix, which represents the local curvature around the current point and refines the gradient direction to speed up convergence. To solve problem \eqref{min_problem1}, Newton's method update at iteration $(k + 1)$ can be written as
\begin{align}
    \label{second_order_update}
    \bm{\theta}^{k + 1} = \bm{\theta}^{k} - \left(\sum_{n=1}^{N}\nabla^{2}f_n(\bm{\theta}^{k})\right)^{-1}\left(\sum_{n=1}^{N}\nabla f_n(\bm{\theta}^{k})\right),
\end{align}
where $\nabla^{2}f_n(\bm{\theta}^{k})\in \mathbb{R}^{d \times d}$ is the local Hessian matrix. For ease of notation, we let $\bm{H}_n^k$ and $\bm{g}_n^k$ represent the Hessian matrix $\nabla^{2}f_n(\bm{\theta}^{k})$ and the gradient vector $\nabla f_n(\bm{\theta}^{k})$ at the $k$-th iteration, respectively. To execute the update specified in \eqref{second_order_update}, every client $n$ computes its local gradient and Hessian. Subsequently, this information is exchanged with the PS to facilitate the model update, and this process continues until convergence. However, the update specified in \eqref{second_order_update} necessitates the transmission of raw Hessians, leading to a communication bottleneck given the substantial size of the Hessian matrix. Moreover, for larger models such as neural networks (NNs), computing and storing the full Hessian is infeasible due to its quadratic memory complexity and significant computational overhead. Furthermore, sharing raw Hessians and gradients unveils significant information about the dataset, rendering them susceptible to inverse attacks and thereby raising privacy concerns. For instance, in linear regression problems, the Hessian matrix is simply the Gramian matrix of the dataset \cite{zhang2024lp}. This disclosure risks exposing sensitive information and may also lead to convergence issues, such as reaching a local maximum.

\section{Proposed Algorithm}\label{proposed_algorithm}
\subsection{Fed-Sophia}
When the model is a NN and $f_n(\bm{\theta})$ is a cross-entropy loss function, using the Gauss-Newton decomposition for the loss function evaluated at the input-label pair $(\bm{x}, \bm{y})$, we can approximate $\bm{H}_n^{k}$ as
\begin{align}
    \label{Hessian_esimate}
     \bm{H}_n^{k} \approx \bm{J}_{\bm{\theta}_n^k}\bm{\phi}(\bm{\theta}_n^k, \bm{x}) \cdot \bm{S} \cdot \bm{J}_{\bm{\theta}_n^k}\bm{\phi}(\bm{\theta}_n^k, \bm{x})^T,
\end{align}
where $\bm{\phi}$ is the logits function or the mapping from input to raw output values, $\bm{J}_{\bm{\theta}_n^k}\bm{\phi}(\bm{\theta}_n^k, \bm{x})$ is the Jacobian of $\bm{\phi}$ with respect to (w.r.t) $\bm{\theta}_n^k$, and $\bm{S}$ is the second-order derivative of the loss w.r.t the logits. 

The diagonal elements of $\hat{\bm{h}}_{n}^k \triangleq \text{diag}(\bm{H}_n^k)$ can be efficiently estimated using the Gauss-Newton-Bartlett (GNB) estimator \cite[Algorithm 2]{liu2023sophia}. Using the diagonal elements of the Hessian is a cost-effective and memory-efficient way to retain curvature information across each dimension while also preserving privacy. Hence, the diagonal elements $\hat{\bm{h}}_{n}^k$ can be expressed as $\hat{\bm{h}}_{n}^k = B \cdot \hat{\bm{g}}_n^k \odot \hat{\bm{g}}_n^k$,
%\begin{align}
%    \hat{\bm{h}}_{n}^k = B \cdot \hat{\bm{g}}_n^k \odot \hat{\bm{g}}_n^k,
%\end{align}
where $\odot$ denotes the element-wise product, $B$ is the mini-batch size, and $\hat{\bm{g}}_n^k$ is the gradient of the mini-batch loss with labels $\hat{\bm{y}}_j$ sampled from the logits function $\bm{\phi}(\bm{\theta}_n^k, \bm{x}_j)$ on the inputs $\bm{x}_j$.

To account for the additional computation overhead, we only compute the diagonal of the Hessian every $\tau$ iterations. Furthermore, we reduce the computation by using mini-batches to estimate the gradient and Hessian. This introduces noise, which can be mitigated using an exponential moving average (EMA) with the following updates
\begin{align}\label{ema_m}
    \bm{m}_{n}^k = \beta_1 \bm{m}_{n}^{k-1} + (1 - \beta_1) \hat{\bm{g}}_n^k,
\end{align}
and for the Hessian 
\begin{align}
\label{ema_h}
\bm{h}_{n}^k = \begin{cases}
    \beta_2 \bm{h}_{n}^{k-1} + (1 - \beta_2) \hat{\bm{h}}_{n}^k, & k \ \text{mod} \ \tau = 0 \\
    \bm{h}_{n}^{k-1}, & \text{otherwise}
\end{cases}
\end{align}
where $\beta_1$ and $\beta_2$ are tuning parameters. The PS aggregates the received EMA of the gradients and the Hessians as $\bar{\bm{m}}^k = \frac{1}{N}\sum_{n=1}^N \bm{m}_{n}^k$, and $\bar{\bm{h}}^{k} = \frac{1}{N}\sum_{n=1}^N \bm{h}_{n}^{k}$, respectively.
%\begin{align}
%    \bar{\bm{m}}^k = \frac{1}{N}\sum_{n=1}^N \bm{m}_{n}^k, \quad \quad %\bar{\bm{h}}^{k} = \frac{1}{N}\sum_{n=1}^N \bm{h}_{n}^{k}.
%\end{align}
Using these EMAs, the PS computes the update direction by performing the element-wise division $\bar{\bm{m}}^k/\bar{\bm{h}}^{k}$. Employing the update direction $\bar{\bm{m}}^k/\bar{\bm{h}}^{k}$ with non-convex functions can result in convergence to a maximum or saddle point, as the Hessian matrix may exhibit rapid changes and inaccuracies, particularly in the early stages of optimization, rendering it unreliable. To address this, we use element-wise clipping to consider values from $\bar{\bm{h}}^{k}$ only when curvature is positive and limit the step size when curvature is steep. Defining the clipping function for a threshold \(\gamma > 0\) of vector \(\bm{z} \in \mathbb{R}^d\) as \(\widehat{\bm{z}} = \text{clip}(\bm{z}, \gamma)\), where \(\widehat{z}_i = \max(\min(z_i, \gamma), -\gamma)\), the update in \eqref{second_order_update} can be written as
\begin{align}
    \label{eq:fed_sophia_update}
     \bm{\theta}^{k+1} = \bm{\theta}^{k} - \eta \cdot \text{clip}\left(\frac{\bar{\bm{m}}^k}{\max\{\gamma \cdot \bar{\bm{h}}^k, \epsilon\}}, 1\right).
\end{align}
Here, \(\epsilon > 0\) ensures positive curvature and avoids division by zero. In \eqref{eq:fed_sophia_update}, very small or negative $\bar{\bm{h}}^k$ entries reduce the pre-conditioned gradient to \(\bar{\bm{m}} / \epsilon\). This triggers clipping, making $\bm{\theta}^{k} - \bm{\theta}^{k + 1}= \eta \cdot \text{sign}(\bar{\bm{m}})$, protecting against worst-case updates when the Hessian estimate is small or negative.
In the subsequent section, we develop the OTA aggregation scheme to further enhance bandwidth utilization.

\subsection{Analog Over-The-Air Direction Aggregation Fed-Sophia (OTA Fed-Sophia)}
\label{OTA_aggregation}
At each iteration $k$, the $N$ clients send their computed EMAs $\bm{m}_n^k$ and $\bm{h}_n^k$ through a wireless fading MAC, through $b$ subcarriers over a total of $\Tilde{T}\in \{T, 2T\}$ time slots, where $b\leq d$ and $T = \lceil \frac{d}{b}\rceil$. To ensure the signals reach the PS simultaneously and utilize the channel's superposition property, clients perform slot-level synchronization, by using, e.g., timing advance in LTE, which adjusts transmission timing to mitigate misalignment and synchronize signal arrival at the PS \cite{etsi36136}. At each time slot $t$ of iteration $k$, each client $n$ transmits a $b$-length vector through the wireless channel denoted by $\bm{S}_{n}(t) = [S_{n, 1}(t), 
 \dots, S_{n, b}(t)]^T \in \mathbb{C}^{b}$. The channel output received by the PS over the $i$th subcarrier is given by, 
\begin{align}
    y_i(t) = \sum_{n=1}^{N} h_{n, i}(t) s_{n, i}(t) + z_i(t),
\end{align}
where $h_{n, i}(t) \in \mathbb{C}$ is the fading channel of the $i$th subcarrier between the PS and client $n$, $z_{i}(t)\in \mathbb{C}$ is the additive white Gaussian noise (AWGN) at the PS at time slot $t$ and follows $\mathcal{C}\mathcal{N}(0, 1)$. In order to facilitate the updates in \eqref{ema_m} and \eqref{ema_h}, for every $\bar{\bm{m}}_i^{k}$ element in $\bar{\bm{m}}^{k}$ the PS needs to perform $\bar{\bm{m}}_{i}^{k} \!=\! \frac{1}{N}\sum_{n=1}^{N}\!\bm{m}_{n,i}^{k}\!+ \!\Re\{z_{i}^{k}\}$, 
%\begin{align}
%    \label{awgn_m_update2}
%    \bar{\bm{m}}_{i}^{k} \!=\! \frac{1}{N}\sum_{n=1}^{N}\!\bm{m}_{n,i}^{k}\!+ \!\Re\{z_{i}^{k}\}, 
%\end{align}
where $\Re(\cdot)$ is the real part function, which can be enabled by  mitigating the effect of the communication channel while satisfying the average transmit power constraint for each client during every transmission round, as expressed by $\frac{1}{d} \sum_{i=1}^{d}|s_{n, i}(t)|^{2} \leq P_n,$
%\begin{align}
%    \label{power_constraint}
%    \frac{1}{d} \sum_{i=1}^{d}|s_{n, i}(t)|^{2} \leq P_n,
%\end{align}
where $P_n$ is the maximum power of client $n$. With the assumption that at each time slot $t$, the channel state information (CSI) is known for the PS and the clients, each client $n$ can utilize the knowledge of $h_{n, i}(t)$ and perform the channel inverse operation. However, to avoid the violation of the power constraint imposed on each client we set the transmitted signal $s_{n, i}(t)$ as
\begin{align}
    \label{transmission_decision}
    s_{n, i}(t) &=  
        \begin{cases}
          \alpha(t) \frac{\bm{m}_{n, i}^{k}}{h_{n, i}(t)}& \text{if } |h_{n, i}(t)| \geq h_{\text{th}}\\
          0& \text{otherwise,}
        \end{cases}
\end{align}
where $\alpha(t)$ represents the scaling factor to be computed at the PS, while $h_{\text{th}}$ denotes a predefined threshold. The expression \eqref{transmission_decision} shows that when the channel condition for client $n$ and subchannel $i$ at time slot $t$ becomes bad, i.e, $|h_{n, i}(t)| < h_{\text{th}}$, the client avoids the transmission and thus not all model parameters are sent to the PS. Thus, we denote the set of elements that can be transmitted, which satisfies the rule in \eqref{transmission_decision}, as $e_n(t) = \{i \in [d]: |h_{n, i}(t)| \geq h_{\text{th}}\}, \forall n \in [N].$
\begin{algorithm}[h]
    \caption{OTA Fed-Sophia}
    \label{alg:_otafedsophia}
    \begin{algorithmic}[1]
        \STATE {\bf Parameters:} $\eta$, hyperparameters $(\beta_1, \beta_2, \epsilon, \gamma)$
        \STATE {\bf Initialize:} $\bm{\theta}^0$, $\bm{m}_n^0 = 0$, $\bm{h}_n^{0} = 0$
        \FOR{communication round $k = 0$ to $K$}
            \FOR{client $n \in [N]$}
                \STATE Receive global model $\bm{\theta}^k$ and set local models
                \STATE Compute local stochastic gradient $\hat{\bm{g}}_n^k$ and update $\bm{m}_{n}^k$
                \STATE Compute the scaling factor $\alpha_n(t)$ 
                and $\alpha(t)$ from\\ \eqref{alpha_power_Constraint} and \eqref{PS_scaling_factor}
                \FOR{$i = 1, ..., d$}
                    \IF{$|h_{n, i}(t)| \geq h_{\text{th}}$}
                    \STATE Send $\alpha(t) \bm{m}_{n, i}^{k}/h_{n, i}(t)$ to the PS
                    \ENDIF
                \ENDFOR
                \IF{$k \mod \tau = 0$}
                    \STATE Compute local Hessian estimate $\bm{\hat{h}}_{n}^k$ and update $\bm{h}_{n}^k$
                    \STATE Compute the scaling factor $\alpha_n(t)$ 
                and $\alpha(t)$ from \eqref{alpha_power_Constraint} and \eqref{PS_scaling_factor}
                \FOR{$i = 1, ..., d$}
                    \IF{$|h_{n, i}(t)| \geq h_{\text{th}}$}
                    \STATE Send $\alpha(t) \bm{h}_{n, i}^{k}/h_{n, i}(t)$ to the PS
                    \ENDIF
                \ENDFOR
                    \STATE Send the updated local Hessian $\bm{\hat{h}}_{n}^k$ to the PS
                \ELSE
                    \STATE $\bm{h}_{n}^k = \bm{h}_{n}^{k-1}$
                \ENDIF
            \ENDFOR
            \STATE Update $\bar{\bm{m}}_i$ and $\bar{\bm{h}}_i^{k}=\sum_{n\in N_i(t)} \bm{h}_{n, i}^{k} + \hat{z}_i^{k}, \forall i \in [d]$ at the PS 
            \STATE Update the global model $\bm{\theta}^{k+1}$ using \eqref{eq:fed_sophia_update}
        \ENDFOR
    \end{algorithmic}
\end{algorithm}

\begin{figure*}[t]
\centering
  \begin{subfigure}{0.24\textwidth}
    \centering
    \includegraphics[width=\textwidth]{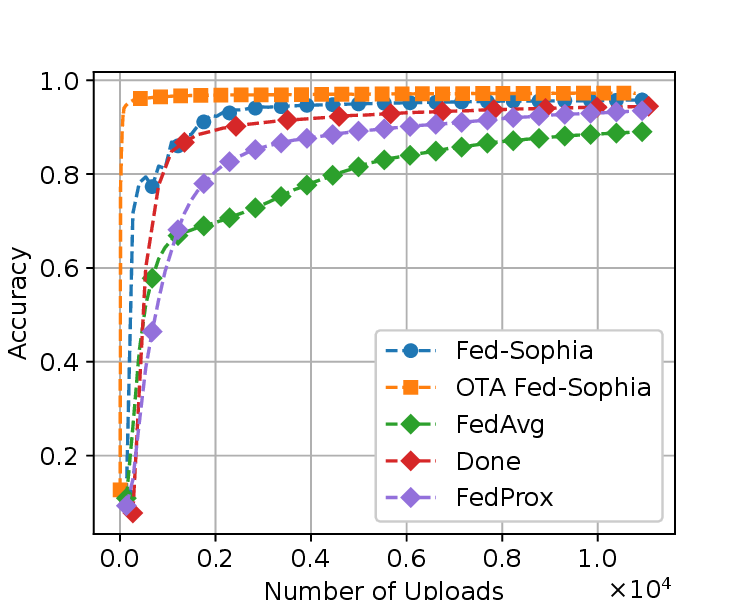}
    \caption{}
    \label{fig_mnist_acc}
  \end{subfigure}
  \begin{subfigure}{0.26\textwidth}
    \centering
    \hspace{-6mm}
    \includegraphics[width=0.85\textwidth, trim=5 5 0 0, clip ]{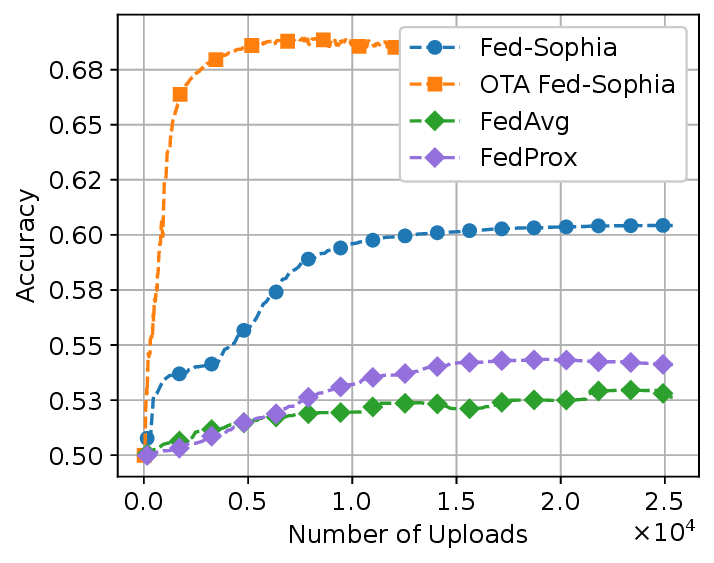}
    \caption{}
    \label{fig_sent140_lstm}
  \end{subfigure}
  \begin{subfigure}{0.24\textwidth}
  
    \centering
    \includegraphics[width=\textwidth]{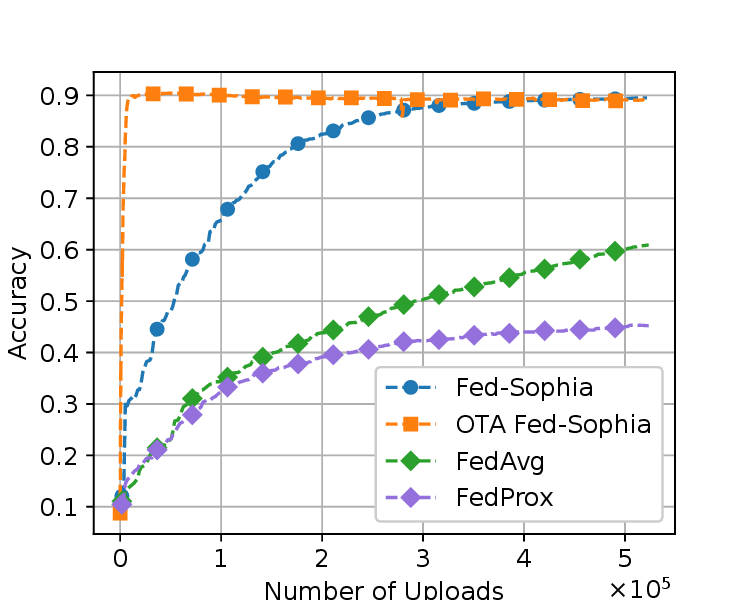}
    \caption{}
    \label{fig_cifar10_acc}
  \end{subfigure}
  \begin{subfigure}{0.24\textwidth}
    \centering
    \includegraphics[width=\textwidth]{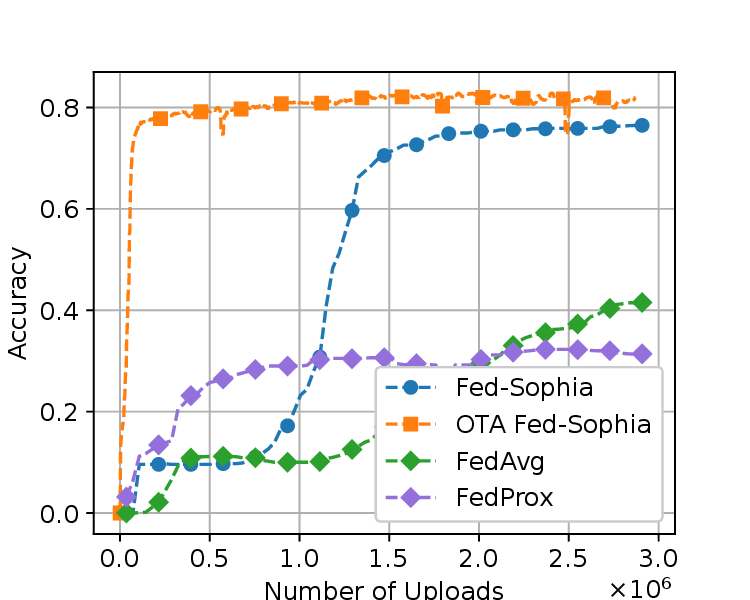}
    \caption{}
    \label{fig_cifar100_acc}
  \end{subfigure}
  \caption{Test accuracy for Fed-Sophia and OTA Fed-Sophia against other baselines in terms of the number of communication uploads for (a) MNIST dataset using MLP model, (b) Sent140 using LSTM model, (c) CIFAR-10 dataset using CNN model, and (d) Cifar-100 using ResNet architecture subfigure (d).}
  \label{fig-1}
\end{figure*}

Since both the PS and the edge clients have the CSI knowledge, every client $n$ can calculate the scaling factor $\alpha_n(t)$ that satisfies the power constraint as follows
\begin{align}
    \label{alpha_power_Constraint}
    \frac{\alpha_n(t)^{2}}{|e_n(t)|} \sum_{i = 1}^{d_n(t)}\left| \frac{\bm{m}_{n, i}^{k}(t)}{h_{n, i}(t)}\right|^{2} \leq P_{n}.
\end{align}
Furthermore, every client $n$ sends its calculated scaling factor $\alpha_n(t)$ to the PS through an error-free channel, and the latter finds the minimum scaling factor satisfying all constraints as
\begin{align}
    \label{PS_scaling_factor}
    \alpha(t) = \text{min}\{\alpha_i(t): i \in e_n(t)\}.
\end{align}
After the global scaling factor $\alpha(t)$ is determined by the PS, it is broadcast to all clients through a control channel. Hence, every client transmits $s_{n, i}(t)$ according to \eqref{transmission_decision} over the $i$th subcarrier and the PS receives the $\alpha(t)\sum_{n\in N_i(t)} \bm{m}_{n, i}^{k}(t) + z_i^{k}(t)$ where $N_i(t) = \{n \in [N]: |h_{n, i}(t)| \geq h_{\text{th}}\}, \quad \forall i \in [d],$ represents the set of clients that can transmits over the $i$th subcarrier. 

Subsequently, the PS divides the expression $\alpha(t)\sum_{n\in N_i(t)} \bm{m}_{n, i}^{k}(t) + z_i^{k}(t)$ by $\alpha(t)$ and applies matched filtering to the received signal, and the processed received signal becomes $\bar{\bm{m}}_i=\sum_{n\in N_i(t)} \bm{m}_{n, i}^{k} + \hat{z}_i^{k}$, where $\hat{z}_i^{k}$ is the matched filtered AWGN. Similarly, after each $\tau$ communication round, the PS obtains the new approximate Hessian $\bar{\bm{h}}_i^{k}=\sum_{n\in N_i(t)} \bm{h}_{n, i}^{k} + \hat{z}_i^{k}$. Finally, the parameter server determines the update direction according to \eqref{eq:fed_sophia_update} and broadcasts the resulting vector to the clients via an error-free communication channel to update the local models. The entire process is provided in Algorithm~\ref{alg:_otafedsophia}.
Although $\bm{m}_i^k$ is recovered exactly, AWGN introduces an error when $\alpha(t)$ is small, amplifying the noise. Jointly optimizing the transmission power and scaling factor to minimize $\sum_{n=1}^N (\bm{m}_n^k - \bar{\bm{m}}^k)$ is the right approach, which is left for future work due to its complexity.

%Subsequently, the PS divides the expression $\alpha(t)\sum_{n\in N_i(t)} \bm{m}_{n, i}^{k}(t) + z_i^{k}(t)$ by $\alpha(t)$ and applies matched filtering to the received signal, and the processed received signal becomes $\bar{\bm{m}}_i=\sum_{n\in N_i(t)} \bm{m}_{n, i}^{k} + \hat{z}_i^{k}$, where $\hat{z}_i^{k}$ is the matched filtered AWGN. Similarly, after each $\tau$ communication round, the PS obtains the new approximate Hessian $\bm{h}_{n, i}(t)$ and applies matched filtering to construct $\bar{\bm{h}}_i^{k}=\sum_{n\in N_i(t)} \bm{h}_{n, i}^{k} + \hat{z}_i^{k}$. Finally, the parameter server determines the update direction according to \eqref{eq:fed_sophia_update} and broadcasts the resulting vector to the edge clients via an error-free communication channel to update the local model of the clients. The entire process is provided in Algorithm~\ref{alg:_otafedsophia}. 

\section{Numerical Results}\label{Simulation_section} 
\subsection{Training Settings}
We consider a classification problem and conduct several experiments to compare the performance of OTA Fed-Sophia with three baselines: FedAvg, FedProx\cite{li2020federated}, and DONE. To demonstrate the performance, we use MNIST, Sent140, CIFAR-10, and CIFAR-100 datasets, employing a multilayer perceptron (MLP), an LSTM, a convolutional NN (CNN), and a ResNet architecture \cite{he2016deep}, respectively. The data is distributed among $N=32$ clients in an independent and identically distributed (IID) manner, with $75\%$ for training and $25\%$ for testing. The cross-entropy loss function is used to train all models.
Each communication round involves 1 local iteration for Fed-Sophia and 10 for FedAvg and FedProx, while DONE requires additional local iterations due to its reliance on multiple updates for convergence. Fed-Sophia, FedAvg, and FedProx use a batch size of 64, whereas DONE processes the full dataset to compute the gradient. To evaluate robustness under non-IID conditions, we also conduct an experiment where data heterogeneity is introduced by limiting each client to a maximum of three labels.
%\vspace{-4mm}
\subsection{Communication Network Settings}
For each client, the transmission power is set to $P_n = 1\text{mW}, \forall n \in [N]$, the available channel bandwidth to $W_{\text{ch}} = 20 \text{MHz}$, and the total number of subcarriers to $b = 1200$. 
Each subcarrier has a bandwidth of $W_{\text{sub}} = 15 \text{KHz}$, with a symbol duration $\tau_{\text{sym}} = 1 \text{ms}$, aligning with LTE standards. The channels' coefficients between clients and the PS are assumed to follow a Rayleigh distribution with zero mean and unit variance $\mathcal{CN}(0, 1)$, and the signal-to-noise ratio is $\text{SNR} = 25 \text{ dB}$.

For OTA {\ours}, analog communication is utilized in the uplink to alleviate the communication bottleneck, while digital communication is employed in the downlink to broadcast the updated model to all clients. Since each element of the update vector is on a subcarrier through a noisy channel on the same subcarrier, all clients need to transmit the entire vector over $\lceil \frac{d}{b}\rceil = \{67, 144, 352, 9350\}$ time slots for the MNIST dataset with an MLP model, Sent140 with an LSTM model, Cifar-10 with a CNN model, and Cifar-100 with a ResNet model, respectively. For the digital {\ours}, DONE, and FedAvg, clients transmit the elements of the update directions using $32$-bit representation, and the required number of transmissions becomes $32d$, and therefore the total number of time slots $\tau_n$ needed by the $n$th client to transmit the entire update vector to the PS, and the total number of available subcarriers is divided among all $N$ clients, is calculated as follows
\begin{align}
    \label{number_tranmsmitted_bits}
    \sum_{s=1}^{\frac{N_s}{N}}\sum_{t=0}^{\tau_n} \tau R_{s, n}(t) \geq 32d,
\end{align}
where $R_{s, n}(t) = BW \, \text{log}_2(1 + P_n |h_{s, n}(t)|^2 / (N_0 \, BW))$ is maximum achievable bit rate at the $t$th time slot, and $N_0$ is the power spectral density of the noise which we set to $10^{-9} \text{W/Hz}$.

The $n$th device's total energy expenditure throughout the training is composed of parts: (i) the communication component, $E_t$, which represents the energy necessary for transmitting and receiving data between the client and the PS, and (ii) the computation component, $E_c$, i.e., the energy required to power the hardware, such as CPUs, GPUs, and memory while performing the computation. After $k$ global iterations, the total energy consumed by device $n$ can be expressed as $E_{n, T}(k)\!=\!E_{n, c}(k)\!+\!E_{n, t}(k)\!=\! \sum_{j=1}^k e_{n}^{j} 
    \!+\!32 d \sum_{j=1}^k  e_{n, PS}^j$,
%\begin{align}
%    E_{n, T}(k)\!=\!E_{n, c}(k)\!+\!E_{n, t}(k)\!=\! \sum_{j=1}^k e_{n}^{j} 
%    \!+\!32 d \sum_{j=1}^k  e_{n, PS}^j,
%\end{align}
where $e_{n}^{j}$ is the energy consumed by $n$th client to train and update the model during $j'$th communication round, and $e_{n, PS}^j$ is the energy needed to transmit one bit to the PS, assuming a 32-bit representation.
%\vspace{-4mm}
\subsection{Performance Comparison} 
In this subsection, we compare the performance of OTA Fed-Sophia against Fed. Sophia, FedAvg, FedProx, and DONE in terms of test accuracy on both the MNIST, Sent140, Cifar-10, and CIFAR-100 datasets w.r.t the number of uploads, as computed in \eqref{number_tranmsmitted_bits}. 

In Figure \ref{fig_mnist_acc}, OTA Fed-Sophia shows the lowest communication uploads, even early on when Hessian estimates are unreliable, thanks to its clipping operation addressing inaccuracies. Despite using a better estimate of the Hessian, DONE requires performing two communication rounds to compute the global model gradient using the entire dataset and still needs to estimate the inverse of the Hessian. For instance, to reach a target test accuracy of 80\%, Fed-Sophia, and OTA Fed-Sophia reach the target accuracy with 80\% fewer uploads compared to the fastest baseline. In fact, Fed-Sophia and OTA Fed-Sophia require $8.1 \times 10^{2}$ and $1.6 \times 10^{2}$ uploads, respectively, while DONE, FedProx, and FedAvg require $9.4 \times 10^{2}$, $2.0 \times 10^{3}$, and $4.6 \times 10^{3}$ uploads, respectively. 

In Figure \ref{fig_sent140_lstm}, we employ the LSTM model on the Sent140 dataset to perform a sentiment analysis task, excluding DONE due to its divergence even with an increased number of local iterations. The results indicate that first-order baselines struggle to match the performance of OTA Fed-Sophia, highlighting the challenges of this problem. Conversely, Fed-Sophia achieves 60\% accuracy within $10^{4}$ less uploads compared to the OTA version.

For the classification task on the CIFAR-10 dataset, Figure \ref{fig_cifar10_acc} illustrates the test accuracy against the number of communication uploads. Fed-Sophia reaches 60\% accuracy within $8 \times 10^{4}$ uploads, which is $6\times$ faster than FedAvg, requiring $4.9 \times 10^{5}$ uploads. In the OTA scenario, Fed-Sophia only requires $4.1 \times 10^{3}$ uploads, yielding over two orders of magnitude speedup compared to FedAvg and FedProx.

Figure \ref{fig_cifar100_acc} further demonstrates the scalability of OTA Fed-Sophia by evaluating ResNet architectures, which contain more than $11.2 \times 10^{6}$ parameters for a classification task on the CIFAR-100 dataset. While other algorithms struggle and show slow convergence, OTA Fed-Sophia only requires $10^{6}$ uploads to achieve 80\% accuracy, and Fed-Sophia reaches 60\% with $1.25 \times 10^{6}$, whereas other baselines remain below 50\% accuracy even with double the communication uploads.
\begin{table}[t]
    \centering
    \caption{Comparison of maximum accuracy for IID and non-IID data distributions after $1.5 \times 10^4$ communication uploads.}
    \begin{tabular}{@{}lccccc@{}}
        \toprule
        & OTA Fed-Sophia & Fed-Sophia & FedAvg & DONE & FedProx \\ 
        \midrule
        IID  & 0.973 & 0.958 & 0.890 & 0.942 & 0.936 \\
        non-IID & 0.870 & 0.858 & 0.621 & 0.835 & 0.862 \\
        \bottomrule
    \end{tabular}
    \label{tab:accuracy_comparison}
\end{table}

In the case of non-IID settings OTA Fed-Sophia demonstrates robust performance by maintaining the highest accuracy level as depicted in Table \ref{tab:accuracy_comparison}. In contrast, FedAvg suffers from significant degradation due to model drift, as client updates diverge during local training to minimize communication costs. FedProx mitigates this issue by penalizing deviations from the global model, stabilizing performance in non-IID conditions. OTA FedSophia addresses client deviation by avoiding multiple local updates and compensates for the extra communication overhead through fast convergence and the efficient use of over-the-air aggregation.
%\vspace{-4mm}
\subsection{Computation Time and Energy Comparison}
FedAvg, FedProx, and (OTA)Fed-Sophia exhibit a linear time complexity of $\mathcal{O}(d)$ per client per iteration, while DONE has a cubic complexity of $\mathcal{O}(d^3)$. Despite this, FedAvg and FedProx require more time and energy to achieve the same accuracy. (OTA)Fed-Sophia's efficiency arises from the GNB estimator, which uses gradients to estimate diagonal elements of the Hessian.
In Table~\ref{tab:cpu_time}, we present the CPU time and emitted carbon footprint needed to reach an $80\%$ accuracy for different algorithms. Our approach outperforms DONE by reaching the target within $85\%$ less CPU time and FedAvg/FedProx by $3 \times$ less. FedAvg, with the lowest computational complexity, exhibits notably slow convergence, while DONE, despite its high complexity per communication round, requires fewer rounds than FedAvg. However, transmitting updates digitally introduces significant delays, especially for large models. Leveraging the superposition property of wireless channels, OTA Fed-Sophia drastically reduces transmission delays, achieving the target accuracy in one-tenth of the CPU time required by the digital approach. Additionally, Fed-Sophia and OTA Fed-Sophia boast the lowest energy consumption, with DONE consuming the most due to its complex calculations. Overall, Fed-Sophia and OTA Fed-Sophia are the most energy-efficient, consuming only $15\%$ of FedAvg's energy, $20\%$ of FedProx's,  and $6.7\%$ of DONE's, with the substantial communication resources OTA Fed-Sophia saves compared to the digital version.

\begin{table}[t]
\centering
\caption{CPU time and carbon footprint required to reach an accuracy of $80\%$ for the different algorithms for the MNIST dataset using an MLP model.}
\begin{tabular}{@{}lcc@{}}
\toprule
\textbf{Algorithm} & \textbf{CPU time (s)} & \textbf{Carbon Footprint (g-CO2-eq)} \\ \midrule
FedAvg & $7.52$ & $1.84E-3$\\
FedProx & $5.37$ & $1.31E-3$\\
DONE & $3.26$ & $4.04E-3$\\
Fed-Sophia & $1.76$ & $0.27E-3$\\
OTA Fed-Sophia & $0.17$ & $0.26E-3$ \\ \bottomrule
\end{tabular}
\label{tab:cpu_time}
\end{table}

\bibliographystyle{IEEEtran}
\bibliography{main.bib} 

\end{document}